# Sigmoid Loss for Language Image Pre-Training

Xiaohua Zhai[⋆]  Basil Mustafa  Alexander Kolesnikov  Lucas Beyer[⋆]
Google DeepMind, Zürich, Switzerland
{xzhai, basilm, akolesnikov, lbeyer}@google.com

## Abstract

*We propose a simple pairwise Sigmoid loss for Language-Image Pre-training (SigLIP). Unlike standard contrastive learning with softmax normalization, the sigmoid loss operates solely on image-text pairs and does not require a global view of the pairwise similarities for normalization. The sigmoid loss simultaneously allows further scaling up the batch size, while also performing better at smaller batch sizes. Combined with Locked-image Tuning, with only four TPUv4 chips, we train a SigLiT model that achieves 84.5% ImageNet zero-shot accuracy in two days. The disentanglement of the batch size from the loss further allows us to study the impact of examples vs pairs and negative to positive ratio. Finally, we push the batch size to the extreme, up to one million, and find that the benefits of growing batch size quickly diminish, with a more reasonable batch size of 32 k being sufficient. We release our models at https://github.com/google-research/big_vision and hope our research motivates further explorations in improving the quality and efficiency of language-image pre-training.*

Table 1: **SigLiT and SigLIP results**. Sigmoid loss is memory efficient, allows larger batch sizes (BS) that unlocks language image pre-training with a small number of chips. SigLiT model with a *frozen public* ❄ B/8 checkpoint [42], trained on the LiT image-text dataset [59] using four TPU-v4 chips for one day, achieves 79.7% 0-shot accuracy on ImageNet. The same setup with a g/14 checkpoint [58] leads to 84.5% accuracy, trained for two days. With a *public unlocked* 🔓 B/16 image checkpoint [42], trained on the WebLI dataset [13], SigLIP achieves 71.0% 0-shot accuracy using 16 TPU-v4 chips for three days. The last two rows show results with randomly initialized models.

|        | Image  | Text | BS   | #TPUv4 | Days | INet-0 |
|--------|--------|------|------|--------|------|--------|
| SigLiT | ❄ B/8  | L[*] | 32 k | 4      | 1    | 79.8   |
| SigLiT | ❄ g/14 | L    | 20 k | 4      | 2    | 84.5   |
| SigLIP | 🔓 B/16 | B    | 16 k | 16     | 3    | 71.0   |
| SigLIP | B/16   | B    | 32 k | 32     | 2    | 72.1   |
| SigLIP | B/16   | B    | 32 k | 32     | 5    | 73.4   |

[*] *We use a variant of the L model with 12 layers.*

## 1. Introduction

Contrastive pre-training using weak supervision from image-text pairs found on the web is becoming the go-to method for obtaining generic computer vision backbones, slowly replacing pre-training on large labelled multi-class datasets. The high-level idea is to simultaneously learn an aligned representation space for images and texts using paired data. Seminal works CLIP [36] and ALIGN [23] established the viability of this approach at a large scale, and following their success, many large image-text datasets became available privately [59, 13, 21, 49] and publicly [40, 6, 15, 7, 41].

The standard recipe to pre-train such models leverages the image-text contrastive objective. It aligns the image and text embeddings for matching (positive) image-text pairs while making sure that unrelated (negative) image-text pairs are dissimilar in the embedding space. This is achieved via a batch-level softmax-based contrastive loss, applied twice to normalize the pairwise similarity scores across all images, then all texts. A naive implementation of the softmax is numerically unstable; it is usually stabilized by subtracting the maximum input value before applying the softmax [18], which requires another pass over the full batch.

In this paper, we propose a simpler alternative: the sigmoid loss. It does not require any operation across the full batch and hence greatly simplifies the distributed loss implementation and boosts efficiency. Additionally, it conceptually decouples the batch size from the definition of the task. We compare the proposed sigmoid loss with the standard softmax loss across multiple setups. In particular, we investigate sigmoid-based loss with two promi-

---

[⋆]equal contribution

nent approaches for image-text learning: CLIP [36] and LiT [59], which we call sigmoid language image pre-training (*SigLIP*) and sigmoid LiT (*SigLiT*), respectively. We find that the sigmoid loss performs significantly better than the softmax loss when the batch size is smaller than 16 k. As the train batch size grows, the gap closes. Importantly, the sigmoid loss is symmetric, requires just a single pass, and a typical implementation requires less memory than the softmax loss. This enables successful training of a SigLiT model at a batch size of *one million*. However, we find that the performance saturates with growing batch size, both for softmax and sigmoid. The good news is that a reasonable batch size, i.e. 32 k, is sufficient for image-text pre-training. This conclusion also holds for multilingual SigLIP training on over 100 languages.

In Table 1, we present setups for image-text pre-training that require a moderate amount of TPUv4 chips for training. SigLiT is surprisingly efficient, reaching 79.7% zero-shot accuracy on ImageNet in just a single day on four chips. SigLIP's more demanding from-scratch training reaches 73.4% zero-shot accuracy in 5 days with 32 TPUv4 chips. This compares favorably to prior works such as FLIP [30] and CLIP [36], which require approximately 5 and 10 days respectively on 256 TPUv3 cores. When fine-tuning a pre-trained vision backbone in SigLIP, denoted as 🔓 in Table 1, we found that disabling the weight decay on the pre-trained backbone leads to better results (see Figure 4 for details). We hope our work paves the way for making the nascent language-image pre-training field more accessible.

## 2. Related Work

**Contrastive learning with the sigmoid loss.** One prior work proposes a similar sigmoid loss for the task of unsupervised dimensionality reduction [19]; in the scope of contrastive image-text learning, the vast majority of works rely on the softmax-based InfoNCE loss as popularized by [46]. In supervised classification, the sigmoid loss has already been shown to be slightly more effective and robust than the softmax loss [3, 51].

**Contrastive language-image pre-training** has become popular since CLIP [36] and ALIGN [23] applied softmax contrastive learning [60, 46, 10, 24] to large-scale image-text datasets. Both models perform very well on zero-shot transfer tasks, including classification and retrieval. Follow-up works show that contrastively pre-trained models produce good representations for fine-tuning [53, 16], linear regression [23], object detection [31], semantic segmentation [33] and video tasks [57].

**Generative language-image pre-training** Besides softmax contrastive pre-training, various alternatives have been proposed. GIT [49], SimVLM [50], and LEMON [21] successfully pre-train models using a generative text decoder instead, while CoCa [56] adds such a decoder to the discriminative CLIP/ALIGN setup, thus combining the pros and cons of both approaches into a single very capable model. BLIP [28] further proposes CapFilt which uses the generative decoder to create better captions and the discriminative part of the model to filter pairs. Language-Image pre-training is a very active field and surveys [8] rapidly become outdated.

**Efficient language-image pre-training** On the other hand, few works have tried making language image pre-training more efficient. LiT [59] and FLIP [30] are notable attempts, the former requires a pre-trained and locked backbone, and the latter sacrifices quality by randomly dropping visual tokens. BASIC [35] and LAION [52] look at scaling batch-size but only go up to 16 k and 160 k respectively, by using many hundreds of chips, and for the former also mixing in a large private classification dataset [35, 55]. The recent Lion optimizer [12] claims to be able to reduce the training cost to reach similar quality.

## 3. Method

In this section, we first review the widely-used softmax-based contrastive loss. We then introduce the pairwise sigmoid loss and discuss its efficient implementation.

Given a mini-batch $\mathcal{B} = \{(I_1, T_1), (I_2, T_2), \dots\}$ of image-text pairs, the contrastive learning objective encourages embeddings of matching pairs $(I_i, T_i)$ to align with each other, while pushing embeddings of unmatched pairs $(I_i, T_{j \neq i})$ apart. For practical purposes, it is assumed that for all images $i$, the text associated with a different image $j$ is not related to $i$, and vice-versa. This assumption is usually noisy and imperfect.

### 3.1. Softmax loss for language image pre-training

When using the softmax loss to formalize this objective, an image model $f(\cdot)$ and a text model $g(\cdot)$ are trained to

**Algorithm 1** Sigmoid loss pseudo-implementation.

```
# img_emb     : image model embedding [n, dim]
# txt_emb     : text model embedding [n, dim]
# t_prime, b  : learnable temperature and bias
# n           : mini-batch size

t = exp(t_prime)
zimg = l2_normalize(img_emb)
ztxt = l2_normalize(txt_emb)
logits = dot(zimg, ztxt.T) * t + b
labels = 2 * eye(n) - ones(n) # -1 with diagonal 1
l = -sum(log_sigmoid(labels * logits)) / n
```

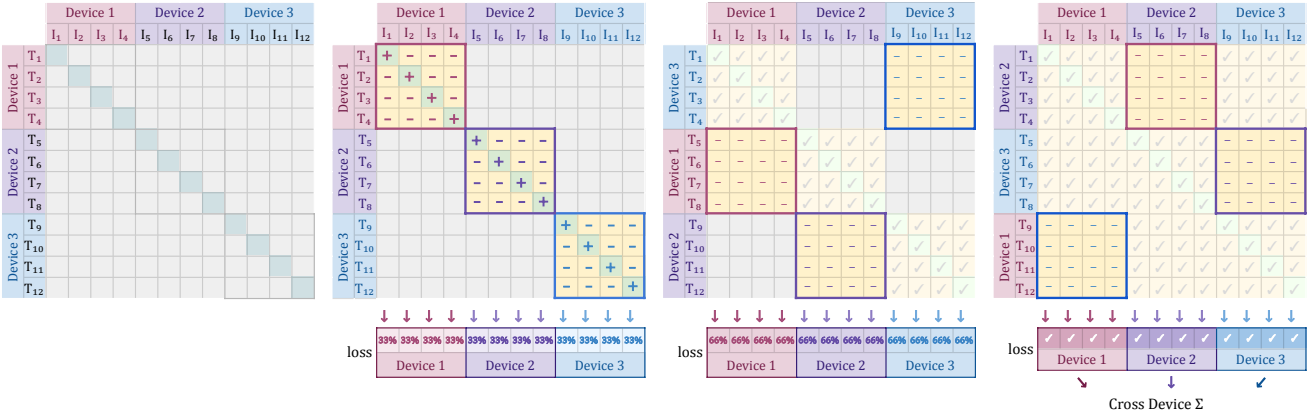

(a) Initially each device holds 4 image and 4 text representations. Each device needs to see the representations from other devices to calculate the full loss.

(b) They each compute the component of the loss (highlighted) for their representations, which includes the positives.

(c) Texts are swapped across the devices, so device 1 now has $I_{1:4}$ and $T_{5:8}$ etc. The new loss is computed and accumulated with the previous.

(d) This repeats till every image & text pair have interacted, e.g. device 1 has the loss of $I_{1:4}$ and $T_{1:12}$. A final cross-device sum brings everything together.

Figure 1: **Efficient loss implementation** demonstrated via a mock setup with 3 devices and a global batch size of 12. There are no all-gathers, and at any point in time only the bright yellow square (size $4 \times 4$) is materialized in memory.

minimize the following objective:

$$-\frac{1}{2|\mathcal{B}|} \sum_{i=1}^{|\mathcal{B}|} \left( \overbrace{\log \frac{e^{t\mathbf{x}_i \cdot \mathbf{y}_i}}{\sum_{j=1}^{|\mathcal{B}|} e^{t\mathbf{x}_i \cdot \mathbf{y}_j}}}^{\text{image}\to\text{text softmax}} + \overbrace{\log \frac{e^{t\mathbf{x}_i \cdot \mathbf{y}_i}}{\sum_{j=1}^{|\mathcal{B}|} e^{t\mathbf{x}_j \cdot \mathbf{y}_i}}}^{\text{text}\to\text{image softmax}} \right)$$

where $\mathbf{x}_i = \frac{f(I_i)}{\|f(I_i)\|_2}$ and $\mathbf{y}_i = \frac{g(T_i)}{\|g(T_i)\|_2}$. In this paper, we adopt the vision transformer architecture [17] for images and the transformer architecture [47] for texts. Note that due to the asymmetry of the softmax loss, the normalization is independently performed two times: across images and across texts [36]. The scalar $t$ is parametrized as $\exp(t')$, where $t'$ is a global freely learnable parameter.

### 3.2. Sigmoid loss for language image pre-training

Instead of the softmax-based contrastive loss, we propose a simpler alternative that does not require computing global normalization factors. The sigmoid-based loss processes every image-text pair independently, effectively turning the learning problem into the standard binary classification on the dataset of all pair combinations, with a positive labels for the matching pairs $(I_i, T_i)$ and negative labels for all other pairs $(I_i, T_{j \neq i})$. It is defined as follows:

$$-\frac{1}{|\mathcal{B}|} \sum_{i=1}^{|\mathcal{B}|} \sum_{j=1}^{|\mathcal{B}|} \underbrace{\log \frac{1}{1+e^{z_{ij}(-t\mathbf{x}_i \cdot \mathbf{y}_j + b)}}}_{\mathcal{L}_{ij}}$$

where $z_{ij}$ is the label for a given image and text input, which equals 1 if they are paired and $-1$ otherwise. At initialization, the heavy imbalance coming from the many negatives dominates the loss, leading to large initial optimization steps attempting to correct this bias. To alleviate this, we introduce an additional learnable bias term $b$ similar to the temperature $t$. We initialize $t'$ and $b$ to $\log 10$ and $-10$ respectively. This makes sure the training starts roughly close to the prior and does not require massive over-correction. Algorithm 1 presents a pseudocode implementation of the proposed sigmoid loss for language image pre-training.

### 3.3. Efficient "chunked" implementation

Contrastive training typically utilizes data parallelism. Computing the loss when data is split across $D$ devices necessitates gathering all embeddings [59] with expensive all-gathers and, more importantly, the materialization of a memory-intensive $|\mathcal{B}| \times |\mathcal{B}|$ matrix of pairwise similarities.

The sigmoid loss, however, is particularly amenable to a memory efficient, fast, and numerically stable implementation that ameliorates both these issues. Denoting the per-device batch size as $b = \frac{|\mathcal{B}|}{D}$, the loss is reformulated as:

$$-\frac{1}{|\mathcal{B}|} \underbrace{\sum_{d_i=1}^{D}}_{\textbf{A: } \forall \text{ device } d_i} \overbrace{\sum_{d_j=1}^{D}}^{\textbf{B: swap negs across devices}} \overbrace{\sum_{i=bd_i}^{b(d_i+1)} \sum_{j=bd_j}^{b(d_j+1)} \mathcal{L}_{ij}}^{\textbf{C: per device loss}}$$

(all local positives / negs from next device)

This is particularly simple for the sigmoid loss as each pair is an independent term in the loss. Figure 1 illustrates this method. In words, we first compute the component of the loss corresponding to the positive pairs, and $b - 1$ negative pairs. We then permute representations across devices, so each device takes negatives from its neighbouring device (next iteration of sum **B**). The loss is then calculated with respect to this chunk (sum **C**). This is done independently in each device, such that each device computes the loss with respect to its local batch $b$. Losses can then simply be summed across all devices (sum **A**). Individual collective permutes (for sum **B**) are fast (and indeed $D$ collective permutes is typically faster than two all-gathers between $D$ devices), and the memory cost at any given moment is reduced from $|\mathcal{B}|^2$ to $b^2$ (for sum **C**). Usually $b$ is constant as scaling $|\mathcal{B}|$ is achieved by increasing the number of accelerators. Due to being quadratic with respect to the batch size, the vanilla loss computation rapidly bottlenecks scaling up. This chunked approach enabled training with batch sizes over 1 million on relatively few devices.

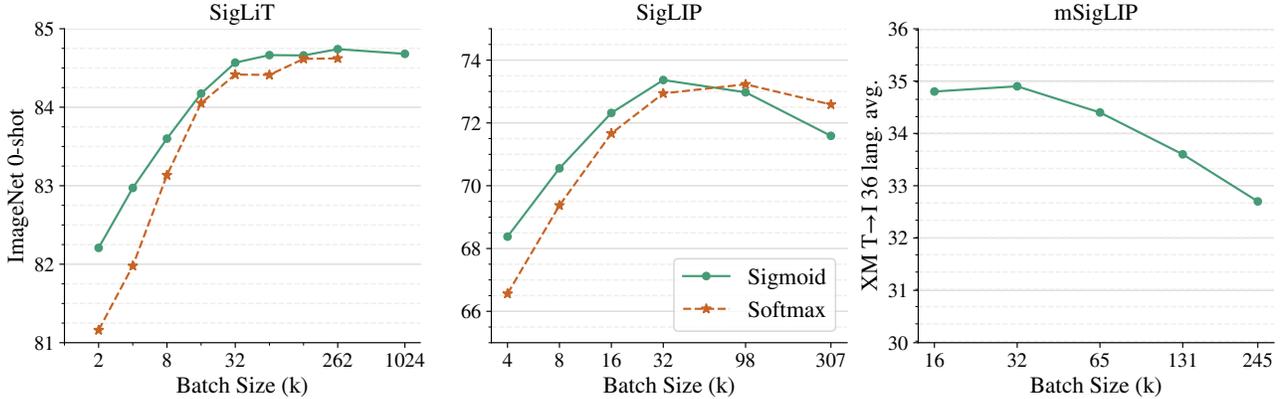

Figure 2: The effect of pre-training batch size. **Left: SigLiT results**, trained for 18B seen examples. Sigmoid loss outperforms the softmax loss significantly with small batch sizes, and performs similarly at larger batch sizes. We successfully trained an SigLiT model with up to *one million* batch size. However, performance for both sigmoid and softmax saturate at around 32 k batch size. **Middle: SigLIP results**, trained for 9B seen examples. Both sigmoid loss and softmax loss saturate at a reasonable batch size, while the peak of the sigmoid loss comes earlier and slightly outperforms the peak of the softmax loss. A very large batch size hurts both losses. **Right: mSigLIP results**, trained for 30B seen examples. With a multilingual setup using over 100 languages, 32 k batch size is surprisingly sufficient and scaling beyond that hurts performance on a 36-language cross-modal retrieval task.

## 4. Results

In this section, we evaluate the proposed SigLiT and SigLIP models across a wide range of batch sizes. We discuss what can be achieved with a small number of accelerator chips, using both SigLiT and SigLIP recipes. We also briefly discuss the impact of batch size on multilingual language image pre-training. We ablate the importance of our large-batch stabilization modification and the introduced learned bias term and present a study on the effect of positive and negative pairs ratio in the sigmoid loss. Lastly, we explore SigLIP's data noise robustness.

To validate our models, we report zero-shot transfer results on the ImageNet dataset [14] and zero-shot retrieval results across 36 languages on the XM3600 dataset [44]. We use the ScalingViT-Adafactor optimizer [58] by default for all our experiments.

### 4.1. SigLiT: Scaling batch size to the limit

Following [59], we use the same precomputed embeddings for the images using a ViT-g vision model, and train a base size text tower from scratch with the same hyperparameters using the LiT image-text dataset [59].

We perform a study over a wide range of batch sizes, from 512 to $1\,M$, demonstrating the impact of batch size for contrastive learning. Results are presented in Figure 2 (left). When the batch size is smaller than $16\,k$, sigmoid loss outperforms softmax loss by a large margin. With growing batch sizes, we observe that softmax loss quickly catches up and potentially slightly underperforms sigmoid loss with a large enough batch size. Overall, we recommend using the SigLIP recipe for large batch sizes as well, due to the simplicity, compute savings, and straightforward memory efficient implementation.

There is a consensus that contrastive learning benefits from large batch sizes, while most of the existing studies stop at 64 k batch size [59, 35, 10]. We successfully trained an SigLiT model at one million batch size, to explore the limit of contrastive learning. To our surprise, the performance saturates at 32 k batch size, further scaling up the batch size only gives a minor boost, and the model peaks at

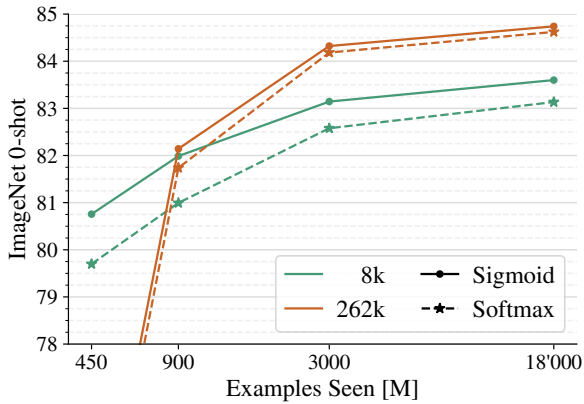

Figure 3: **SigLiT ImageNet 0-shot transfer results with different training durations.** Large batch size results in a big performance boost, but needs a sufficiently long schedule to ramp up, as for short schedules, very large batch size results in a small number of gradient update steps.

|       | 16 k | 32 k | 64 k | 128 k | 240 k |
|-------|------|------|------|-------|-------|
| INet-0 | 71.6 | 73.2 | 73.2 | 73.2 | 73.1 |
| XM avg | 34.8 | 34.9 | 34.4 | 33.6 | 32.7 |
| XM de | 54.7 | 54.8 | 55.4 | 54.3 | 54.7 |
| XM en | 46.5 | 46.2 | 46.5 | 46.6 | 46.6 |
| XM hi | 9.1 | 8.5 | 7.9 | 8.1 | 7.3 |
| XM ru | 50.1 | 49.9 | 49.7 | 48.6 | 49.3 |
| XM zh | 30.7 | 32.5 | 32.0 | 30.6 | 23.7 |

Table 2: Multilingual SigLIP results with various batch sizes, pre-trained for 30 billion seen examples. We report zero-shot transfer results on ImageNet (INet-0) and averaged text to image retrieval results across 36 languages on the crossmodal 3600 dataset (XM). The full table on 36 languages can be found in Appendix.

256 k batch size. Our best SigLiT with a *B*-sized text mode achieves 84.7% zero-shot transfer accuracy on ImageNet, while the original LiT paper reports a slightly better 85.2% score with a 10 times larger *g*-sized text model. Figure 3 presents the impact of training duration for different batch sizes. It demonstrates that large, $262\,k$ batch size significantly outperforms smaller $8\,k$ batch size when trained for a sufficiently long time. Note, that for short training durations, large batch size leads to the fewer absolute number of update steps and thus needs more time to ramp up.

### 4.2. SigLIP: Sigmoid loss is beneficial for language-image pre-training

We pre-train SigLIP models on the WebLI dataset [13], using only English image and text pairs. We use CLIP (WebLI) to denote the CLIP baseline pre-trained on WebLI with the standard softmax loss. We use moderately-sized models: B/16 ViT for image embeddings and B-sized transformer for text embeddings. The input images are resized to 224×224 resolution. The text is tokenized by a 32 k vocabulary sentencepiece tokenizer [27] trained on the English C4 dataset [37], and a maximum of 16 text tokens are kept. Figure 2 middle plot shows SigLIP results, With less than 32 k batch size, SigLIP outperforms CLIP (WebLI) baselines. On the other end of the scale, the memory efficiency of the sigmoid loss enabled much larger batch sizes. For example, with four TPU-v4 chips, we could fit a batch size of 4096 with a Base SigLIP but only 2048 with a corresponding CLIP model. The two advantages together demonstrate significant benefits of the sigmoid loss for language image pre-training with fixed resources, which will be discussed in Section 4.5.

As batch size increases, the gap between the sigmoid and the softmax losses diminish. SigLIP performs best at batch size 32 k, whereas the softmax loss required 98 k for optimal performance and still didn't outperform the sigmoid based variant. Scaling further, a larger batch size like 307 k hurts both losses.

### 4.3. mSigLIP: Multi-lingual pre-training

We further scale up the training data by keeping all the *100 languages* from the WebLI dataset [13]. With multilingual data, one usually needs to use a larger international vocabulary. We first verify the impact of two tokenizers: a small multilingual vocabulary with 32 k tokens [37], and a large multilingual vocabulary with 250 k tokens [54]. We train B-sized ViT and text models for $900\,M$ total examples seen, and observe slightly more than 1% improvement when using a larger vocabulary.

However, the token embeddings become huge for very large vocabulary sizes. Following the standard setup, we would need to store a $N \times W$ token embedding lookup table to train the multilingual model, where $N$ is the vocabulary size mentioned above and $W$ is the embedding dimension of the text model. To save memory, we propose to use a "bottlenecked" token embedding. We use $N \times K$ embedding matrix and additional $K \times W$ projection, where the bottleneck $K$ is much smaller than $W$.

In our experiments, we observed that using a large multilingual vocabulary with a bottleneck can be scaled up as efficiently as using a small multilingual vocabulary. Specifically, by enabling the bottleneck of size $K = 96$ for Base architecture with $W = 768$, we only see about a half percent quality drop on ImageNet zero-shot transfer, compared to using the full $250k$ vocabulary.

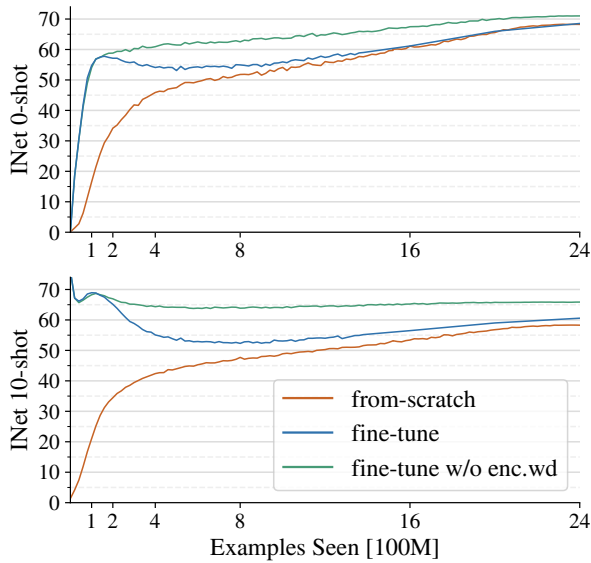
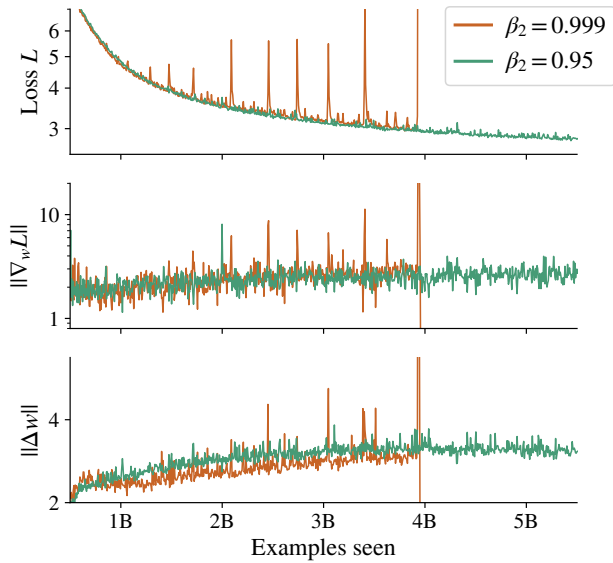

Figure 4: **Top**: SigLIP with pre-trained encoders ramps up quickly. However, only disabling weight decay on the pre-trained encoder weights leads to stable behavior and good ImageNet 0-shot transfer results. **Bottom**: ImageNet 10-shot transfer results, where decaying the pre-trained weights leads to deterioration of the pre-trained model visual representation quality. Disabling weight decay flattens the curve.

Figure 5: **The effect of Adam and AdaFactor's $\beta_2$.** As we increase batch-size, we observe more frequent training instability. This instability seen in the loss curves (top) is caused by spikes in gradient norm (middle) leading to large parameter updates (bottom). Decreasing the $\beta_2$ momentum stabilizes training. Occasional gradient spikes still happen (see step at 2B), but do not destabilize the training process.

With the memory improvements, we train mSigLIP models for various batch sizes, for a total of 30 billion examples seen. Table 2 and Figure 2 (right plot) show the results. We were expecting a large batch size to improve multilingual pre-training, where the model sees more examples from the same language as hard negatives in a single mini-batch. However, we didn't observe clear improvements with a batch size larger than 32 k. A batch size of 32 k is sufficient for a multilingual setup as well. On the XM3600 cross-modal retrieval tasks, we found that going beyond 32 k batch size leads to worse results on average while on ImageNet zero-shot transfer it stays flat. mSigLIP sets the new state-of-the-art on XM3600 text to image retrieval task, with only a Base size model. Our best result is 34.9%, which is more than 6% higher than the previously reported result 28.5% [13] with a standard LiT model [59] using a much larger four billion ViT-e model. We further scale up mSigLIP training in Section 4.6.

### 4.4. SigLiT with four TPU-v4 chips

For many practitioners, the important question usually is "what can be trained with a limited amount of resources?" We explore the usage of SigLiT models in this section with only four TPU-v4 chips, as the memory efficient sigmoid loss is suitable for this application scenario.

We follow the same setup as in section 4.1. We use the publicly available ViT-AugReg-B/8 [42] model as the frozen (❄) vision tower, and precompute embeddings to accelerate the training [59]. The text model is a Large Transformer, but with a depth of only 12 layers (instead of 24). It is trained using the LION [12] optimizer with decoupled weight decay $1 \times 10^{-7}$, linearly warm-up of learning rate over 6.5k steps up to a peak of $1 \times 10^{-4}$, followed by a cosine decay to 0. We train for a total of 65 000 steps with a batch size of 32k – this leads to just under one day of training. Table 1 shows the results when training a model on four chips for one day, achieving 79.7% 0-shot ImageNet classification accuracy; very competitive in this limited resource regime. With a ViT-g/14 [58] model as the vision tower and a Large text tower, we can train at 20 k batch size on four chips for 107 k steps in under two days. This further pushes the 0-shot ImageNet classification accuracy up to 84.5%.

### 4.5. SigLIP with a small amount of TPU-v4 chips

It's resource demanding to train a CLIP model from-scratch in general, with SigLIP it's possible to fit a larger train batch size with fewer amount of chips. In this section, we explore ways to train SigLIP models efficiently with pre-trained weights. We use pre-trained weights to initialize the image model to accelerate the pre-training, which was orig-inally discussed in [59]. We use the public and unlocked 🔓 ViT-AugReg-B/16 [42] model to initialize our vision tower and fine-tune on the same WebLI English data as used for SigLIP. In all the experiments, we apply a 0.1 learning rate multiplier to the pre-trained image tower to make it suitable for fine-tuning.

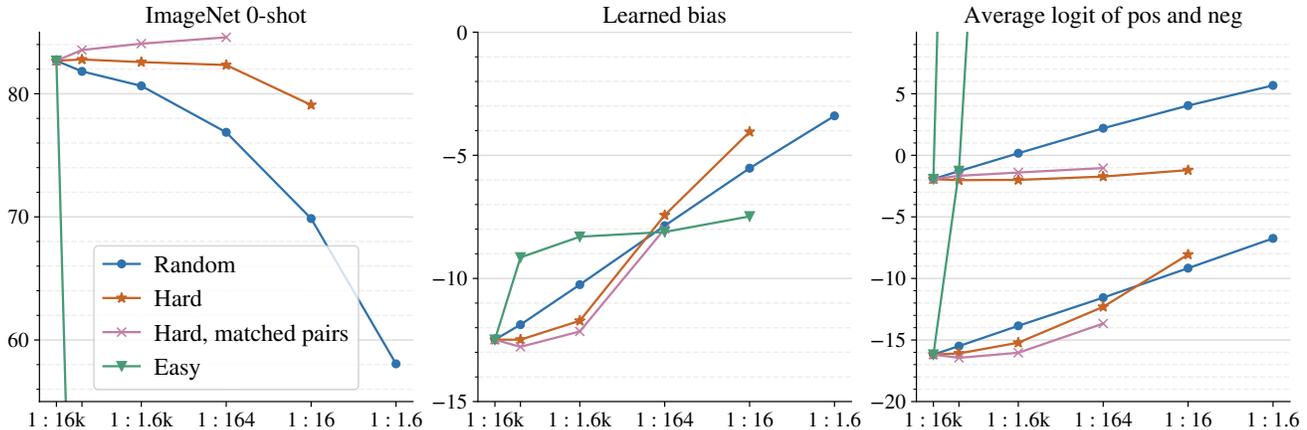

Figure 6: **The effect of batch composition.** We simulate various batch compositions by masking out negatives, either randomly, keeping only the hardest, or the easiest. With no masking, we have 16 k negatives for each positive in the batch (1:16 k) and the strongest masking we apply (1:1.6) results in almost balanced minibatches. In one setting we *match total pairs* seen by training for significantly longer. We observe ImageNet 0-shot score, the final value of the learned bias, and the average logits of positive and negative pairs. Overall, the imbalance does not seem to be detrimental, but finding an *efficient* way of mining negatives might be beneficial.

Figure 4 presents unlocked 🔓 fine-tuning results alongside from-scratch randomly initialized baselines. We used 16 TPU-v4 chips and train at 16 k batch size for 2.4 B examples seen. We found that the fine-tuning setup doesn't perform well out-of-the-box; this is consistent with prior works [59] where finetuning image models degraded visual representation quality. This is evidenced by ImageNet 10-shot linear classification, where in Figure 4 the fine-tuned setup is barely better than the from-scratch baseline.

We hypothesize that the default weight decay applied to the pre-trained weights reduces their effectiveness. Motivated by the fine-tuning recipe from [17, 58, 25], that uses no weight decay, we also propose disabling weight decay on the pre-trained weights for SigLIP training. Weight decay is therefore only applied to the randomly initialized weights in the text model. This simple modification significantly improved SigLIP results. Figure 4 shows that with our improved recipe, SigLIP reaches 71% 0-shot accuracy on ImageNet, using $16k$ batch size, trained on 16 chips for three days. We also present from-scratch results in the bottom rows of Table 1: with 32 TPUv4 chips for only two days, SigLIP achieves 72.1% 0-shot accuracy. This presents a significant training cost reduction e.g. compared to CLIP (approx. 2500 TPUv3-days for 72.6%) reported in [30].

### 4.6. Scaling up SigLIP and mSigLIP

In this section, we scale up SigLIP by "overtraining" the model [45, 1]. We present results in Table 3 using ViT-B, ViT-L or So-400m [1] as the vision encoder, with a text encoder of the same size (B, L and So-400m respectively). Following the recipe described in Section 4.2, we train both models for 40 billion examples seen at batch size 32 k, but use $(256/16)^2 = 256$ image patches and 64 text tokens (instead of 16). To get SigLIP models for different resolutions, we train for 5 billion more examples at the target resolution, with a 100x smaller learning rate and no weight decay. In Table 3, we report zero-shot classification results on ImageNet [14], ObjectNet [2], ImageNet-v2 [39], ImageNet ReaL [3], and zero-shot image-to-text (I→T) retrieval, text-to-image (I→T) retrieval results on MSCOCO [11].

We also scale up the multilingual mSigLIP ViT-B model in the same way. We report image-text retrieval results across 36 languages on the XM3600 benchmark [44]. The scaled-up mSigLIP ViT-B model achieves the state-of-the-art *42.6% image retrieval recall@1 and 54.1% text retrieval recall@1* for a Base model. This is slightly outperformed by the Large model in [48] getting 42.96% image retrieval recall@1. Detailed results are provided in Appendix Table 9 and Figure 8, denoted as *32 k.

### 4.7. Stabilizing large-batch training

As we move to large batch sizes, the language image pretraining using transformers becomes increasingly more unstable, even when using a modestly-sized model (e.g. Base size). The reason for these instabilities is large spikes in the gradient norms, which translate to large-magnitude changes in the weights that may destabilize the training process, see Figure 5. We observe that reducing $\beta_2$ in Adam and AdaFactor from its default 0.999 to 0.95 (which was suggested in [20, 9]) is enough to stabilize the training. Intuitively, this allows recovering from gradient spikes quicker. We opt for setting $\beta_2 = 0.95$ for all our experiments.

| Method | Image Encoder | | ImageNet-1k | | | | COCO R@1 | |
|---|---|---|---|---|---|---|---|---|
| | ViT size | # Patches | Validation | v2 | ReaL | ObjectNet | I→T | T→I |
| CLIP | B | 196 | 68.3 | 61.9 | - | 55.3 | 52.4 | 33.1 |
| OpenCLIP | B | 196 | 70.2 | 62.3 | - | 56.0 | 59.4 | 42.3 |
| EVA-CLIP | B | 196 | 74.7 | 67.0 | - | 62.3 | 58.7 | 42.2 |
| SigLIP | B | 196 | **76.2** | **69.6** | 82.8 | **70.7** | **64.4** | **47.2** |
| SigLIP | B | 256 | 76.7 | 70.0 | 83.1 | 71.3 | 65.1 | 47.4 |
| SigLIP | B | 576 | 78.6 | 72.1 | 84.5 | 73.8 | 67.5 | 49.7 |
| SigLIP | B | 1024 | **79.2** | **73.0** | **84.9** | **74.7** | **67.6** | **50.4** |
| CLIP | L | 256 | 75.5 | 69.0 | - | 69.9 | 56.3 | 36.5 |
| OpenCLIP | L | 256 | 74.0 | 61.1 | - | 66.4 | 62.1 | 46.1 |
| CLIPA-v2 | L | 256 | 79.7 | 72.8 | - | 71.1 | 64.1 | 46.3 |
| EVA-CLIP | L | 256 | 79.8 | 72.9 | - | 75.3 | 63.7 | 47.5 |
| SigLIP | L | 256 | **80.5** | **74.2** | 85.9 | **77.9** | **69.5** | **51.1** |
| CLIP | L | 576 | 76.6 | 72.0 | - | 70.9 | 57.9 | 37.1 |
| CLIPA-v2 | L | 576 | 80.3 | 73.5 | - | 73.1 | 65.5 | 47.2 |
| EVA-CLIP | L | 576 | 80.4 | 73.8 | - | 78.4 | 64.1 | 47.9 |
| SigLIP | L | 576 | **82.1** | **75.9** | 87.0 | **81.0** | **70.6** | **52.7** |
| OpenCLIP | G (2B) | 256 | 80.1 | 73.6 | - | 73.0 | 67.3 | 51.4 |
| CLIPA-v2 | H (630M) | 576 | 81.8 | 75.6 | - | 77.4 | 67.2 | 49.2 |
| EVA-CLIP | E (5B) | 256 | 82.0 | 75.7 | - | 79.6 | 68.8 | 51.1 |
| SigLIP | SO (400M) | 729 | **83.2** | **77.2** | 87.5 | **82.9** | **70.2** | **52.0** |

Table 3: **Comparison with other publicly released models.** Our SigLIP models outperform all prior models, e.g. Open-CLIP [22] and CLIP [36], by a significant margin on both zero-shot classification and retrieval tasks. Compared to the concurrent EVA-CLIP [43] and CLIPA-v2 [29], our SigLIP-L performs better across the board, in both the low and high resolution cases. Especially noteworthy is the Shape-Optimized 400M parameter ViT [1] architecture, which outperforms all significantly larger models. We publicly release our models: https://github.com/google-research/big_vision.

### 4.8. Negative ratio in sigmoid loss

One question which arises when shifting the perspective from the softmax's "pick the right class" view to the sigmoid's "rate this pair" view, is the imbalance in positive versus negative pairs. For a batch size $|\mathcal{B}|$, the batch contains $|\mathcal{B}|$ positive pairs, but $|\mathcal{B}|^2 - |\mathcal{B}|$ negative examples. In the modest batch-size of 16 k, there are actually 268 M negative examples for only 16 k positive ones. At the same time, because the sigmoid loss decomposes into a sum of per-example losses, we can perform controlled experiments to study the effect of the mini-batch composition and dis-tribution of examples visited. We run experiments in the SigLiT setup at batch-size 16 k for 900 M steps and vary the composition of the batch by masking out (*i.e.* ignoring) enough negative examples to reach a target "positive : negative" ratio, masking in the following ways:

- **Random:** Randomly choose negative pairs to mask.
- **Hard:** Keep hardest negative pairs (highest loss).
- **Easy:** Keep easiest negatives pairs (lowest loss).
- **Hard + matching total pairs seen:** Masking examples while training for a fixed number of steps does decrease the total number of *pairs* seen during training. Hence in the *matched pairs* setting, we increase the number of training steps by the masking ratio in order to keep the number of pairs seen constant.

Figure 6 shows the effect of the various masking strategies. Randomly removing negatives to rebalance does deteriorate performance. Keeping the easiest examples does not work at all, while keeping the hardest negatives does almost maintain the quality, indicating that, as could be expected, a lot of the learning on the negative side comes from the harder examples. This is further confirmed by the slightly increased performance of training longer on the hardest examples in order to match the total pairs seen.

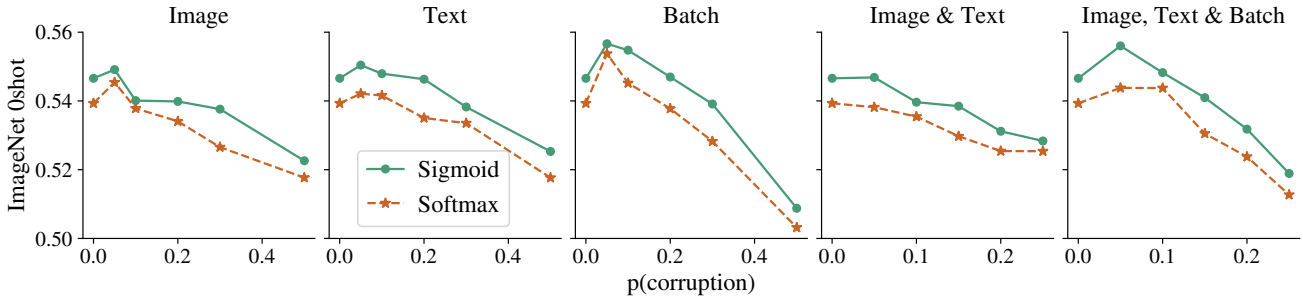

Figure 7: **Sigmoid-training increases robustness** to data noise. Titles show the type of corruption applied, and x-axes show the probability with which they are applied. With increasing corruption severity, M-scale models trained with sigmoid loss for 3.6 billion examples retain superiority over corresponding softmax baseline.

We also look at the value of the learned bias at the end of training as well as the average logit value for positive and negative examples across these settings, and find the result mostly follows what one would expect: as fewer negatives are present, the bias and logits become more positive overall. Interestingly, when training with more hard negative pairs, the average logits of positive pairs stays mostly flat.

This study confirms that (1) the imbalance does not seem to be a major reason for concern, while at the same time (2) coming up with an *efficient* way of including more negative examples can be promising but is not trivial.

### 4.9. Bias term in sigmoid loss

We ablate the bias term in the loss function, using the Base architecture with an 8 k batch size, trained for 900M examples with the SigLIP setup. Zero-shot transfer results are reported on ImageNet [14], Oxford-iiit pet [34] and Cifar100 [26]. Table 4 presents results with and without a bias term in the sigmoid loss.

Table 4: **Bias (b) and temperature (t′) initialization.** Results are reported using Base architecture, 8 k batch size, trained for 900M examples. Enabling the bias term b with $-10$ initialization improves results consistently.

| b | t′ | INet-0 | Pet-0 | C100-0 |
|---|---|---|---|---|
| n/a | log 10 | 62.0 | 81.8 | 59.9 |
| -10 | log 10 | **63.0** | **82.4** | **61.0** |
| -10 | log 1 | 61.0 | 80.0 | 60.4 |
| 0 | log 10 | 61.7 | 79.9 | 59.0 |
| 0 | log 1 | 53.7 | 73.2 | 53.8 |

Enabling the bias term with a $-10$ initialization consistently improves performance across all tasks. This is because the bias term ensures that the training starts close to the prior, preventing dramatic over-correction in early optimization. In contrast, a randomly chosen bias term initialization, such as the 0 initialization in Table 4, fails to address the over-correction issue, leading to significantly worse results. This effect is particularly noticeable when using a small temperature $t'$ initialization. We set the bias and temperature initialization to $b = -10$ and $t' = \log 10$ (hence $t = 10$) as the default for all experiments.

### 4.10. Label noise robustness

Prior works demonstrated improved robustness against label noise when using the sigmoid loss for classification models [3]. This property would be particularly useful here in the face of the famously noisy nature of popular large-scale image-text datasets. In order to study this for SigLIP, we train M/16 image models alongside an M text model at batch size 16384 for 3.6 billion seen examples. We corrupt the training data using one of the following methods:

- **Image**: With probability $p$, replace the image with uniform random noise.
- **Text**: With probability $p$, replace tokenized text with a new sequence of randomly sampled tokens, up to some (sampled) sequence length.
- **Batch alignment**: Randomly shuffle the ordering of $p\%$ of the batch.
- **Image & text**: Apply both with probability $p$ each.
- **Image, text & batch**: Alongside (4), also shuffle fraction $p$ of alignments.

Results from varying the likelihood of the corruption are shown in Figure 7. Models trained with sigmoid loss are increasingly robust to all kinds of added noise.

## 5. Conclusion

We conducted a study on two language-image pre-training instances that used the sigmoid loss: SigLiT and SigLIP. Our results demonstrate that the sigmoid loss performs better than the softmax baseline, particularly for small train batch sizes. This loss function is also more memory efficient, which allows larger train batch sizes without requiring additional resources. We performed a thorough investigation of the batch size in contrastive learning. Surprisingly, we found that a relatively modest batch size of 32 k yielded nearly optimal performance. Further studies have been performed to understand better the introduced bias term in the sigmoid loss, robustness to data noises and the impact of positive and negative pairs ratio in the sigmoid loss. We hope this work will facilitate language-image pre-training research with limited resources.

**Acknowledgements.** We thank Daniel Keysers, Ilya Tolstikhin, Olivier Bousquet and Michael Tschannen for their valuable feedback and discussions on this paper. We thank Joan Puigcerver, Josip Djolonga and Black Hechtman for discussions on efficient implementations of the chunked contrastive loss. We thank Kaiming He and Xinlei Chen for the discussion of $\beta_2$ to stabilize the training. We also thank Ross Wightman for spotting a mistake in the pseudocode in the first version of this paper, Boris Dayma and Krzysztof Maziarz for spotting typos in the second and third versions which made $t$ vs $t'$ confusing. We thank the Google Deepmind team for providing a supportive research environment. We use the `big_vision` codebase [5, 4] for all experiments in this project.

## A. More results for SigLiT

In section 4.1, we use the same precomputed embeddings for the images using a ViT-g vision model from [59]. Only resize augmentation is applied, to a fixed $288 \times 288$ resolution. We train a standard base size text tower, using the ScalingViT-Adafactor optimizer [58] with $\beta_1 = 0.9$ and $\beta_2 = 0.95$. We use 0.001 learning rate with a linear warmup schedule for the first 200 M examples seen, and then the learning rate is decayed to zero with a cosine learning rate schedule. Weight decay is set to 0.0001 for all the experiments. The text is tokenized by a 32 k vocabulary sentencepiece tokenizer [27] trained on the English C4 dataset [37], and a maximum of 16 text tokens are kept. Table 8 shows results with multiple train examples seen and batch sizes, for both the sigmoid loss and the softmax loss baseline.

For training SigLiT in under a day with 4 chips (Section 4.4), we used the LION optimizer with peak learning rate $1 \times 10^{-4}$ and weight decay $1 \times 10^{-7}$. The learning rate was warmed linearly to the peak in 6.5 k steps, then cosine decayed to zero for the remaining 58.5 k steps.

## B. More results for SigLIP

In Table 5, we present more results for SigLIP Base with multiple train examples seen: 3 billion examples and 9 billion examples respectively.

| Batch Size | 3 B | | 9 B | |
|---|---|---|---|---|
| | sigmoid | softmax | sigmoid | softmax |
| 512 | **51.5** | 47.7 | - | - |
| 1 k | **57.3** | 53.2 | - | - |
| 2 k | **62.1** | 59.3 | - | - |
| 4 k | **65.3** | 63.8 | **68.4** | 66.6 |
| 8 k | **68.6** | 66.6 | **70.6** | 69.4 |
| 16 k | - | - | **72.3** | 71.7 |
| 32 k | **69.9** | **69.9** | **73.4** | 72.9 |
| 98 k | 69.5 | **69.7** | 73.0 | **73.2** |
| 307 k | - | - | 71.6 | **72.6** |

Table 5: **SigLIP zeor-shot accuracy (%) on the ImageNet benchmark.** Both the sigmoid loss and the softmax loss baseline are presented. Experiments are performed on multiple train examples seen (3 B, 9 B) and train batch sizes (from 512 to 307 k). When trained for 9 B examples, the peak of the sigmoid loss comes earlier at 32 k than the peak of the softmax loss at 98 k. Together with the memory efficient advantage for the sigmoid loss, it allows one to train the best language-image model with much fewer amount of accelerators.

| BS | Default | Best | Best LR | Best WD |
|---|---|---|---|---|
| 8 k | 70.1 | 70.1 | 0.001 | 0.0001 |
| 16 k | 70.0 | 70.0 | 0.001 | 0.0001 |
| 32 k | 68.2 | 69.0 | 0.0003 | 0.00003 |

Table 6: Default hyperparameters across different batch sizes, perform either the best or close to the best hyperparameter from a sweep. Zero-shot accuracy on ImageNet is reported. BS=batch size, LR=learning rate, WD=weight decay.

## C. Robustness of SigLIP results

**Hyperparameters for different batch sizes.** Sigmoid loss doesn't require tuning hyperparameters for different batch sizes. For example, in both the SigLiP and SigLiT setup, we only used default 0.001 learning rate and 0.0001 weight decay across a wide range of batch sizes (from 512 to 1024k). We further performed a sweep of 9 hyperparameters across 3 batch sizes on the from-scratch SigLIP setup for 3B seen examples: learning rate {0.0003, 0.001, 0.003} × weight decay {0.00003, 0.0001, 0.0003} × batch size {8 k, 16 k, 32 k}. We observed in Table 6 that the default LR/WD is either the best or close to the best.

**Standard deviation.** We repeat SigLIP training five times, using the recommended 32k batch size and 3B seen examples. We report the average and std in Table 7. The std of the five runs is very small for both sigmoid and softmax.

**Alternative optimizers.** We repeat the same experiment with AdamW optimizer five times and got very similar results and std as reported in Table 7. We tested a linear learning rate scheduler instead of the default cosine learning rate scheduler, it achieves 69.9% accuracy.

## D. More results for mSigLIP

We present the mSigLIP Base crossmodal retrieval results on the Crossmodal-3600 dataset, across all the 36 langauges in Figure 8 and Table 9.

| Loss | Optimizer | Results (%) |
|---|---|---|
| Softmax | ViT-Adafactor | $69.9 \pm 0.1$ |
| Sigmoid | ViT-Adafactor | $70.1 \pm 0.2$ |
| Sigmoid | AdamW | $70.3 \pm 0.1$ |

Table 7: Mean and standard deviation of five repeated experiments. Zero-shot accuracy on ImageNet is reported.

| Batch Size | 450 M | | 900 M | | 3 B | | 18 B | |
| --- | --- | --- | --- | --- | --- | --- | --- | --- |
| | sigmoid | softmax | sigmoid | softmax | sigmoid | softmax | sigmoid | softmax |
| 512   | 72.5 | 69.5 | 75.0 | 72.8 | 77.2 | 74.6 | -    | -    |
| 1 k   | 75.5 | 73.6 | 77.2 | 76.0 | 79.6 | 77.9 | -    | -    |
| 2 k   | 77.1 | 76.3 | 79.3 | 78.1 | 81.3 | 80.1 | 82.2 | 81.2 |
| 4 k   | 79.2 | 78.3 | 80.8 | 79.8 | 82.4 | 81.2 | 83.0 | 82.0 |
| 8 k   | 80.8 | 79.7 | 82.0 | 81.0 | 83.1 | 82.6 | 83.6 | 83.1 |
| 16 k  | 81.2 | 81.2 | 82.7 | 82.1 | 83.8 | 83.5 | 84.2 | 84.1 |
| 32 k  | 81.9 | 81.4 | 83.1 | 82.7 | 84.2 | 84.0 | 84.6 | 84.4 |
| 64 k  | 81.6 | 81.6 | 83.0 | 82.8 | 84.3 | 84.1 | 84.7 | 84.4 |
| 128 k | 80.5 | 80.0 | 83.1 | 83.2 | 84.2 | 84.4 | 84.7 | 84.6 |
| 256 k | 72.8 | 72.2 | 82.1 | 81.7 | 84.3 | 84.2 | 84.7 | 84.6 |
| 1024 k| -    | -    | -    | -    | -    | -    | 84.7 | -    |

Table 8: **SigLiT zero-shot accuracy (%) on the ImageNet benchmark.** Both the sigmoid loss and the softmax loss baseline are presented. Extensive experiments are performed on multiple train examples seen (450 M, 900 M, 3 B, 18 B) and train batch sizes (from 512 to 1 M).

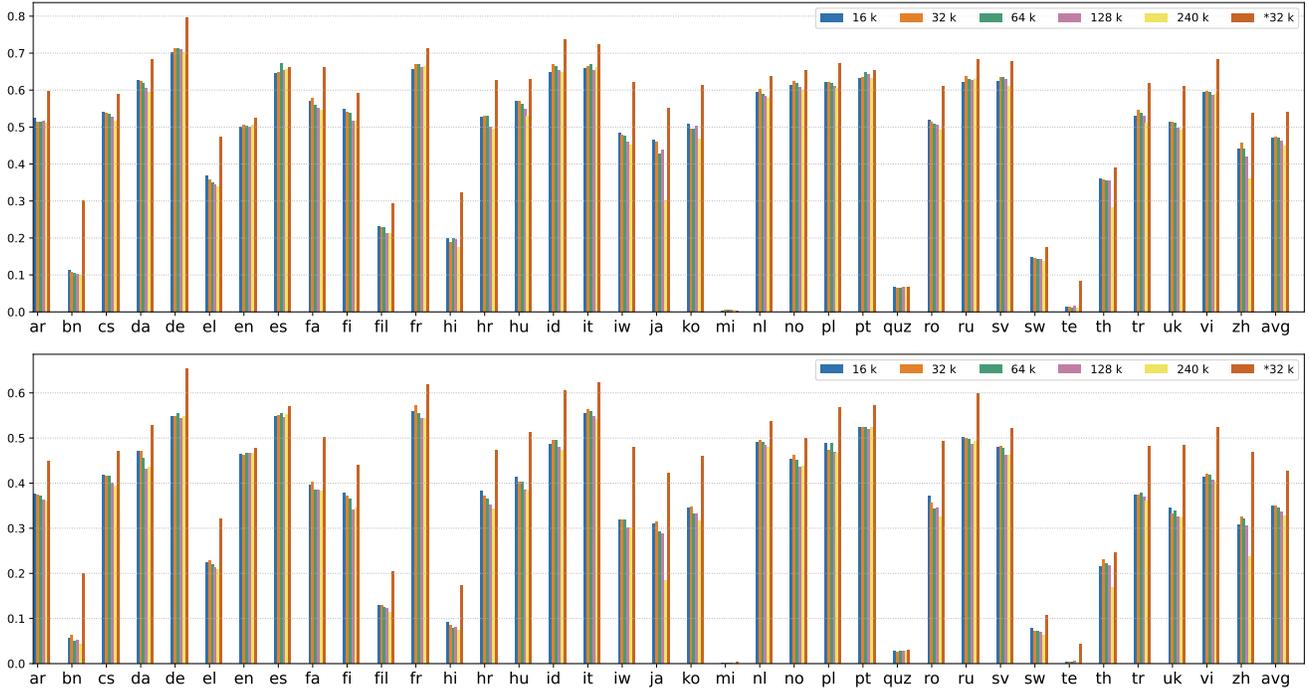

Figure 8: **Image-to-text and text-to-image zero-shot retrieval recall@1 results on all 36 languages of Crossmodal-3600**. Top: Image to text. Bottom: text to image. Colors are batch sizes. *32 k represents the scaled up results as described in Section 4.6.

## E. Label noise experiments

All models had an M/16 image tower and a M text tower. They were trained from random initialisation for 3.6B examples seen, with a batch size of 16384. A cosine learning rate schedule was used, with an initial linear warmup for 10% of steps up to a peak learning rate of 0.001.

| Lang. | Image-to-text | | | | | | Text-to-image | | | | | |
|---|---|---|---|---|---|---|---|---|---|---|---|---|
| | 16 k | 32 k | 64 k | 128 k | 240 k | *32 k | 16 k | 32 k | 64 k | 128 k | 240 k | *32 k |
| ar | 52.4 | 51.3 | 51.5 | 51.5 | 51.1 | 59.7 | 37.6 | 37.4 | 37.1 | 36.3 | 36.0 | 44.9 |
| bn | 11.4 | 10.8 | 10.4 | 10.3 | 9.9 | 30.1 | 5.5 | 6.2 | 4.9 | 5.1 | 4.4 | 20.0 |
| cs | 54.1 | 53.7 | 53.7 | 52.8 | 51.8 | 58.9 | 41.8 | 41.6 | 41.5 | 39.9 | 39.4 | 47.0 |
| da | 62.7 | 62.4 | 62.0 | 60.4 | 59.3 | 68.4 | 47.0 | 47.0 | 45.6 | 43.0 | 43.5 | 52.9 |
| de | 70.3 | 71.4 | 71.2 | 71.1 | 70.2 | 79.7 | 54.7 | 54.8 | 55.4 | 54.3 | 54.7 | 65.3 |
| el | 36.9 | 35.8 | 35.1 | 34.5 | 33.8 | 47.4 | 22.4 | 22.8 | 22.0 | 21.3 | 20.8 | 32.2 |
| en | 50.1 | 50.5 | 50.2 | 49.9 | 50.7 | 52.5 | 46.5 | 46.2 | 46.5 | 46.6 | 46.6 | 47.6 |
| es | 64.7 | 64.9 | 67.2 | 65.3 | 65.6 | 66.3 | 54.8 | 55.0 | 55.5 | 54.5 | 55.2 | 57.0 |
| fa | 57.0 | 57.8 | 56.1 | 55.3 | 54.6 | 66.2 | 39.6 | 40.2 | 38.4 | 38.4 | 38.3 | 50.0 |
| fi | 54.9 | 54.1 | 53.8 | 51.7 | 51.7 | 59.1 | 37.7 | 37.1 | 36.4 | 34.0 | 34.5 | 44.0 |
| fil | 23.2 | 22.8 | 22.9 | 21.4 | 21.2 | 29.2 | 12.8 | 12.9 | 12.4 | 12.2 | 11.3 | 20.4 |
| fr | 65.7 | 66.9 | 67.0 | 66.1 | 66.5 | 71.2 | 55.9 | 57.1 | 55.5 | 54.4 | 54.3 | 61.8 |
| hi | 19.9 | 18.8 | 19.9 | 19.5 | 17.4 | 32.2 | 9.1 | 8.5 | 7.9 | 8.1 | 7.3 | 17.3 |
| hr | 52.7 | 53.0 | 53.0 | 49.9 | 49.6 | 62.6 | 38.2 | 37.1 | 36.4 | 35.2 | 34.3 | 47.2 |
| hu | 57.0 | 57.1 | 56.3 | 54.8 | 53.0 | 62.9 | 41.4 | 40.2 | 40.2 | 38.6 | 38.2 | 51.2 |
| id | 64.8 | 67.1 | 66.6 | 65.4 | 64.7 | 73.7 | 48.5 | 49.4 | 49.5 | 47.8 | 47.3 | 60.5 |
| it | 65.9 | 66.4 | 67.1 | 65.2 | 66.1 | 72.3 | 55.5 | 56.4 | 55.8 | 54.8 | 54.1 | 62.3 |
| iw | 48.4 | 47.9 | 47.7 | 46.1 | 45.2 | 62.2 | 31.8 | 31.8 | 31.9 | 30.1 | 30.1 | 48.0 |
| ja | 46.4 | 45.9 | 42.9 | 43.7 | 30.2 | 55.1 | 31.0 | 31.3 | 29.2 | 28.9 | 18.5 | 42.3 |
| ko | 50.8 | 49.5 | 49.4 | 50.2 | 46.8 | 61.4 | 34.4 | 34.7 | 33.2 | 33.1 | 31.5 | 45.9 |
| mi | 0.4 | 0.4 | 0.6 | 0.6 | 0.4 | 0.3 | 0.2 | 0.2 | 0.2 | 0.2 | 0.2 | 0.3 |
| nl | 59.6 | 60.4 | 58.9 | 58.3 | 57.9 | 63.6 | 48.9 | 49.5 | 48.9 | 48.4 | 47.9 | 53.6 |
| no | 61.4 | 62.4 | 62.0 | 60.9 | 59.9 | 65.3 | 45.3 | 46.2 | 45.0 | 43.5 | 43.7 | 50.0 |
| pl | 62.2 | 62.0 | 62.0 | 61.1 | 60.5 | 67.1 | 48.8 | 47.4 | 48.7 | 46.8 | 46.7 | 56.7 |
| pt | 63.1 | 63.6 | 64.9 | 64.3 | 63.2 | 65.4 | 52.4 | 52.3 | 52.3 | 51.9 | 52.4 | 57.3 |
| quz | 6.8 | 6.4 | 6.4 | 6.6 | 6.7 | 6.8 | 2.7 | 2.6 | 2.7 | 2.7 | 2.8 | 2.9 |
| ro | 52.1 | 51.4 | 51.0 | 50.6 | 49.3 | 61.0 | 37.2 | 35.6 | 34.3 | 34.5 | 32.5 | 49.3 |
| ru | 62.2 | 63.6 | 63.1 | 62.7 | 63.1 | 68.4 | 50.1 | 49.9 | 49.7 | 48.6 | 49.3 | 59.9 |
| sv | 62.3 | 63.5 | 63.5 | 63.1 | 61.2 | 67.7 | 47.9 | 48.2 | 47.6 | 46.2 | 46.2 | 52.0 |
| sw | 14.8 | 14.4 | 14.3 | 14.2 | 13.8 | 17.4 | 7.8 | 7.2 | 7.1 | 6.9 | 6.3 | 10.7 |
| te | 1.2 | 1.2 | 1.2 | 1.7 | 1.1 | 8.4 | 0.4 | 0.3 | 0.3 | 0.5 | 0.3 | 4.3 |
| th | 36.1 | 35.8 | 35.6 | 35.6 | 28.3 | 39.0 | 21.6 | 23.1 | 22.2 | 21.6 | 16.8 | 24.6 |
| tr | 53.1 | 54.5 | 53.7 | 52.9 | 51.2 | 62.0 | 37.3 | 37.4 | 37.8 | 37.0 | 36.1 | 48.1 |
| uk | 51.4 | 51.5 | 51.2 | 49.9 | 49.2 | 61.2 | 34.5 | 33.2 | 33.8 | 32.5 | 32.4 | 48.3 |
| vi | 59.6 | 59.8 | 59.5 | 58.5 | 58.8 | 68.4 | 41.4 | 41.9 | 41.9 | 40.6 | 40.3 | 52.3 |
| zh | 44.1 | 45.7 | 44.1 | 41.9 | 36.1 | 53.9 | 30.7 | 32.5 | 32.0 | 30.6 | 23.7 | 46.8 |
| **avg** | 47.2 | 47.4 | 47.1 | 46.3 | 45.0 | **54.1** | 34.8 | 34.9 | 34.4 | 33.6 | 32.7 | **42.6** |

Table 9: **Image-to-text (text retrieval) and text-to-image (image retrieval) zero-shot recall@1 results on all 36 languages of Crossmodal-3600**, with mSigLIP models trained at different batch sizes for 30 B total examples seen. *32 k represents the scaled up results as described in Section 4.6.

## F. Model Card

We provide a description of our models following [32].

- **Model Architecture:** The model is trained using the contrastive pre-training technique with sigmoid loss as described in this paper. This contrastive model contains two encoders, i.e. vision transformer encoder [17] and language transformer encoder [47]. The vision and language encoders always have the same size, one of ViT-B, ViT-L and SoViT-400M [1].

- **Inputs:** The vision encoder takes an image ($224 \times 224 \times 3$, $256 \times 256 \times 3$, $384 \times 384 \times 3$, $512 \times 512 \times 3$) as input. The text encoder takes a tokenized text [38, 54] cropped to the first 64 tokens as input.

- **Outputs:** The vision and text encoders both output a $d$ dimensional feature vector, where $d$ is 768, 1024 and 1152 for ViT-B, ViT-L and SoViT-400M, respectively.

- **Intended Use:** The models are designed for multi-modal research purposes. The models can be used for zero-shot image classification and zero-shot image-text retrieval by comparing both feature vectors. We provide both en-only and i18n-trained models to encourage research on the impact of this choice.

- **Training Data:** The contrastive model is pre-trained from-scratch using the WebLI [13] dataset. SigLIP models are pre-trained on a WebLI subset filtered to contain mostly English. mSigLIP models are pre-trained on the WebLI dataset without language filters.

- **Evaluation Data:** Zero-shot classification is performed on ImageNet [14], ImageNet v2 [39], ImageNet Real [3], and ObjectNet [2]. Zero-shot retrieval is performed on COCO [11] and the multilingual XM3600 dataset [44].

- **Hardware & Software**: The models are developed in the `big_vision` codebase [5, 4] and trained on Google Cloud TPUs.